\documentclass[letterpaper, 10 pt, conference]{ieeeconf}  

\IEEEoverridecommandlockouts                              

\usepackage{cite}
\usepackage{amsmath,amssymb,amsfonts}
\usepackage{algorithmic}
\usepackage{graphicx}
\usepackage{textcomp}
\usepackage{hyperref}
\usepackage{cleveref}
\usepackage{booktabs}
\usepackage{comment}
\usepackage{tabularx}
\usepackage{nicematrix}
\usepackage{adjustbox}
\usepackage{afterpage}
\usepackage[english]{babel}
\def\BibTeX{{\rm B\kern-.05em{\sc i\kern-.025em b}\kern-.08em
    T\kern-.1667em\lower.7ex\hbox{E}\kern-.125emX}}

\addto\extrasenglish{%
}
\title{\LARGE \bf
The Machine Vision Iceberg Explained: Advancing Dynamic Testing by Considering Holistic Environmental Relations}
\author{Hubert Padusinski$^{1}$,  Christian Steinhauser$^{1}$, Thilo Braun$^{1}$, Lennart Ries$^{1}$, Eric Sax$^{1}$
\thanks{$^{1}$Hubert Padusinski, Christian Steinhauser, Thilo Braun, Lennart Ries and Eric Sax are with FZI Research Center for Information Technology, 76131 Karlsruhe, Germany
        {\tt\small padusinski@fzi.de, steinhauser@fzi.de,  braun@fzi.de, ries@fzi.de, sax@fzi.de}}%
}

\begin{document}
\maketitle
\thispagestyle{empty}
\pagestyle{empty}

\begin{abstract}
Machine Vision (MV) is essential for solving driving automation. This paper examines potential shortcomings in current MV testing strategies for highly automated driving (HAD) systems. We argue for a more comprehensive understanding of the performance factors that must be considered during the MV evaluation process, noting that neglecting these factors can lead to significant risks. This is not only relevant to MV component testing, but also to integration testing. To illustrate this point, we draw an analogy to a ship navigating towards an iceberg to show potential hidden challenges in current MV testing strategies. The main contribution is a novel framework for black-box testing which observes environmental relations. This means it is designed to enhance MV assessments by considering the attributes and surroundings of relevant individual objects. The framework provides the identification of seven general concerns about the object recognition of MV, which are not addressed adequately in established test processes. To detect these deficits based on their performance factors, we propose the use of a taxonomy called "granularity orders" along with a graphical representation. This allows an identification of MV uncertainties across a range of driving scenarios. This approach aims to advance the precision, efficiency, and completeness of testing procedures for MV.

\end{abstract}
\section{Introduction}
\label{sec:introduction}

Several institutions \cite{vvmprojekt} \cite{sakuraproject} \cite{ccamsunriseproject} are integrating HAD systems\cite{SAE2021}  across all traffic domains. The development of Machine Vision (MV) technology is crucial to enable systems to interpret incoming visual information of the ODD. The intention is that HAD systems are progressively deployed in all environments: From restricted properties, like factories, agricultural spaces to all public environments, such as highways, urban or rural environments. Embedded in the HAD cause chain, MV functions must handle an inherent richness of context: Recognition of navigable routes or obstacles and detecting instructions. All these functions must be adaptable to various scenes — from bustling downtown areas to serene green spaces amid challenging weather circumstances.
Currently, researchers are focused on implementing MV using artificial neural network models for detecting and classifying relevant objects
\cite{girshick2014rich}\cite{redmon2016you}\cite{milioto2018rangenet}\cite{wu2018squeezeseg} \cite{alzubaidi2021}. However, past publications have shown that insufficient training, originated by imbalances or knowledge gaps in the training data, can lead to unexpected results \cite{oksuz2020imbalance} \cite{johnson2019survey}. Misclassifications may occur due to specific attributes, such as patterns on T-shirts, or unknown relations, like objects behind unfamiliar backgrounds. These uncertainties of MV over the operating domain can cause intolerable results during deployment leading to harmful accidents. Consequently, test methods are required to proof the overall robustness of an arbitrary test object. This is a significant challenge for organizations as the public demands a freedom of failures, but institutions are limited in development time frames. Our work addresses the challenges of the three common test strategies testing MV, recognizing a call for a holistic confrontation with MVs performance factors. To overcome the iceberg challenge, we present our main contributions:
\begin{figure}
    \centering
    \includegraphics[width=0.9\linewidth]{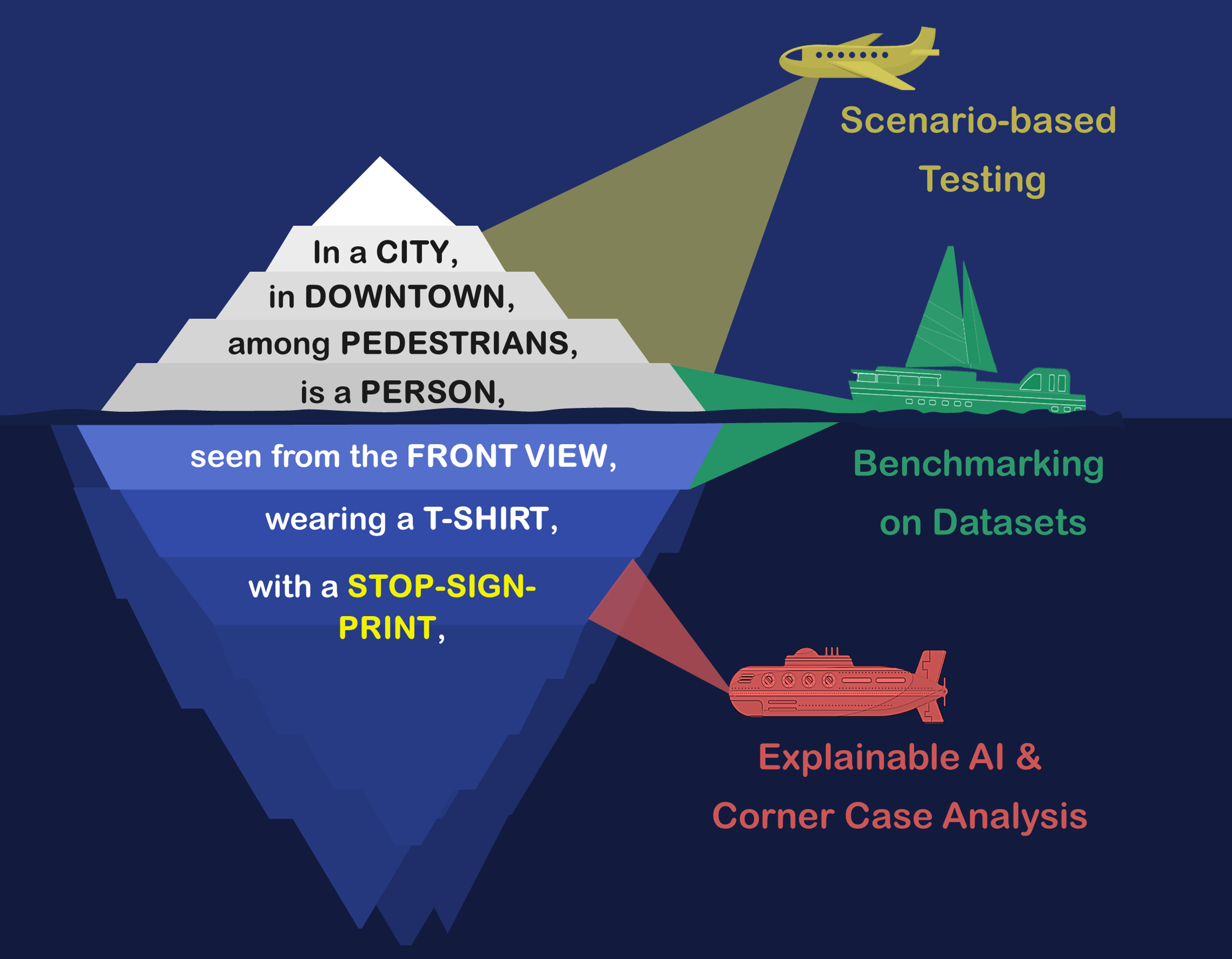}
    \caption{The strategies challenged by overcoming the metaphorical machine vision iceberg. While each can see the iceberg's hurdles from a high range of view, from a straight up view or a deep focus, none are able to see the full extent of the iceberg.}
\label{fig:methapher}
\end{figure}

\begin{itemize}
    \item [-] \textbf{Collection of General Object Recognition Deficits:}
A list of seven fundamental concerns about MV systems that sets a baseline of what should be relevant in their evaluation. Developed from an in-depth comparative analysis of human psychology versus MV deficits.
    \item [-] \textbf{Taxonomy of Granularity Orders:}
A novel taxonomy that radically rethinks the structure of ODDs in their elevations. Allowing a decomposition of entities based on their scaling in environments. Supports the systematizing attributes of relevant objects and their surroundings.
    \item [-] \textbf{Evaluation with Environmental Entity Relation Graph:}
A graphical representation that connects observed environmental entities in their occurrence relation. By classifying and comparing the occurrence of relations based on the reaction of MV systems, the representation enables to argue deficits over a multitude of driving scenarios.
\end{itemize}

\section{Motivation}
\label{sec:motivation}

With Fig. \ref{fig:methapher}, we introduce a metaphorical mental model to demonstrate the challenge of integrating opaque MV functions into context-rich environments: The \textit{Machine Vision Iceberg}. Understanding the nature of an iceberg requires delving beyond its obvious hurdles and grasping its entirety in a holistic manner. Various strategies have emerged to navigate the hurdles posed by the MV Iceberg:

The \textit{straightforward strategy} of benchmarking functions akin to an icebreaker opting for an efficient, but rough path by acknowledging occasional collisions. Developers benchmark against established data sets \cite{Geiger2012CVPR} \cite{Behley2019iccv} to express an average accuracy over their MV functions. But these datasets do not sufficiently represent all required circumstances for a HAD-System, as they contain only randomized and limited distribution of circumstances \cite{petersen22}. \textit{Deep-Dive Strategies}, like explainable AI (XAI) \cite{arrieta2020xai} \cite{breitenstein2020}, analogous to submarines, delve into opaque aspects of the iceberg. Their strength lies in uncovering of fundamental factors rather than offering a holistic evaluation during a development process, but are slower in comparision to the other strategies. The \textit{overlooking strategy} of scenario-based testing (SBT) may seem comprehensive but lacks the nuanced perspective needed for those actively dealing with the hurdles. The analogy of an airplane with a bird's-eye view of the iceberg illustrates this limitation. SBT \cite{winkle2016development} \cite{king2021} \cite{Neurohr2021} \cite{scholtes21} achieves high coverage across the entire system and domains, but falls short in specifying relevant causal understanding crucial for MV.
To approximate a holistic understanding of the MV iceberg, we aim to incorporate the insights of all these strategies. This prompts us to explore the iceberg from tip to a manageable depth of understanding with the key question: \textit{How deeply do testers need to delve into relations of ODDs to sufficiently evaluate the robustness of MV in a development process?}
\section{Current Strategies and their Challenges}
\label{sec:current_challenges}

\subsection{Deep-Dive Strategy: Unraveling MV}
MV systems incorporate sensors, hardware computing units, and neural network-based functions. These functions are characterized as high-dimensional processing chains with non-linear transformations \cite{alzubaidi2021}, so that the integrated models of these systems often appear as black boxes \cite{Buhrmester2019AnalysisOE} to developers and testers.  While corner case analyses (CCA) \cite{breitenstein2020} and XAI \cite{arrieta2020xai} approaches offer efficient general performance statements, they primarily focus on identifying causes of individual malfunctions through resource-intensive analysis of specific cases or extreme values. Traditional test data compilation often relies on explicit failure conditions from prior accidents deemed relevant by humans. This approach can lead to imbalanced data \cite{oksuz2020imbalance}, potentially causing the system to exhibit irrational behavior from human point of view \cite{theverge2016tesla}. The deep-dive strategy provide valuable fundamental understanding of specific conditions requiring analysis within MV systems. Deep dive strategies involve time-consuming investigations, such as inspecting hidden layers of neural networks, interpreting activation functions, or manual inspections. This time-consuming investigation of new deficiencies limits the ability to comprehensively investigate the variety of possible specific circumstances across the operational domain. Due to the required stimulation with highly different domain-, task- and vendor-specific sensor data, their methods cannot be used for a holistic robustness statement over MV test objects.

\textit{Challenge 1: Deep-dive strategies allow to fundamentally decipher uncertainties of MV, but their procedures are too time-consuming to investigate the full extent of interplay between MV and its operating domains during verification and validation of HAD systems.}

\subsection{Straightforward Strategy: Overgeneralized Benchmarking}

Neural networks are commonly evaluated using benchmark datasets \cite{Geiger2012CVPR} \cite{Behley2019iccv}\cite{RobotCarDatasetIJRR}\cite{Cordts2016Cityscapes}. The straightforward strategy evaluating datasets, play a crucial role in an efficient estimation of the performances directly during developing neural networks, fostering innovation in the field. In the context of MV systems, real-world datasets often consist of raw sensor data (camera images, point clouds) captured by vehicle-mounted sensors. However, generating these datasets involves significant manual effort for labeling sensor data with ground truth information. Performance evaluation in object detection typically involves quantifying the deviation between test object results and ground truth masks. Metrics such as Intersection-over-Union (IoU) are used to analyze accuracy over time series \cite{johnson2019survey}. State-of-the-art approaches leverage mean Intersection-over-Union (mIoU) \cite{padilla2020} on specific object classes within the test data to efficiently assess MV function performance \cite{girshick2014rich}\cite{redmon2016you}\cite{milioto2018rangenet}\cite{wu2018squeezeseg}. While this approach provides valuable insights, it can be insufficient for determining the overall robustness of MV systems if the datasets lack sufficient diversity, such as limited variations in vehicle colors \cite{petersen22}. This can lead to overfitting, where trained models excel on specific instances within the training data but fail to generalize to real-world scenarios. For example, a network excelling at recognizing pedestrians within the training data might struggle with recognizing stooped people who are underrepresented in the dataset. Consequently, high mIoU for the "person" class can create a false sense of robustness, as the network performs poorly in specific, real-world situations.

\textit{Challenge 2: Benchmarking is a valuable method for quantifying the generalisability of neural networks across object classes. However, it cannot reveal specific uncertainties about an object due to over averaging performances over object classes within possibly incomplete test data.}

\subsection{Overlooking Strategy: Blind Spots for SBT}

SBT for HAD-Systems is a valuable approach to systematically and dynamically tests automated driving functions under conditions derived from real-world concerns \cite{riedmaier2020} \cite{Neurohr2021}. Scenarios are constructed using real-world drive data \cite{langner2019}, test benches, or simulations, with parameters like velocities and distances varied during execution to create comprehensive test statements. Current research primarily focuses on overall system behavior during maneuvers, considering kinematic parameters and interactions with dynamic objects like Vulnerable Road Users (VRUs) and cars \cite{king2021}. From a MV perspective, the dynamic variation in these test cases primarily involves kinematic states, encompassing positions and scaling of dynamic objects in injected sensor data. That means SBT systematize only a short scope of underlying concerns that MV faces what is known from the knowledge derived with XAI. One major limitation of SBT is its focus on the systematic variation of individual object attributes, such as size, color, and shape. This approach fails to consider the critical role that object relationships play in MV. The mere variation of object attributes within a scene does not adequately reflect the  interactions and relationships that exist between objects in real-world environments. For example, the recognition of a foreground object, like a person can be significantly influenced by the surrounding scene \cite{xiao2020noise}. An MV test object could reliably recognize people in a city, but as soon as it is carried out in an underrepresented district, people cannot be recognized with the associated class. The fundamental 6-layer model for SBT \cite{scholtes21} provides a framework for categorizing the multitude of objects in scenarios based on their functional properties and dynamic behaviour. However, the model does not capture their attributes and their relations to the overall scene, e.g. this lack of relations between dynamic and partly also static objects limits the SBT to consider the robustness of MV systems in scenarios, which is crucial for object recognition. This shows the need for a model that incorporates this depth of information based on known concerns.

\textit{Challenge 3: SBT, which aims to construct test cases with relevant factors for HAD systems, does not take into account the concerns and circumstances relevant for MV. Relations between objects their attributes and their immediate surroundings are blind spots for SBT, which can influence the result of executed scenarios.}

\section{Concept}
\label{sec:concept}

\begin{figure}[h]
    \centering
      \begin{adjustbox}{margin=4ex 4ex,width=0.45\textwidth}
        \includegraphics[width=1.0\textwidth]{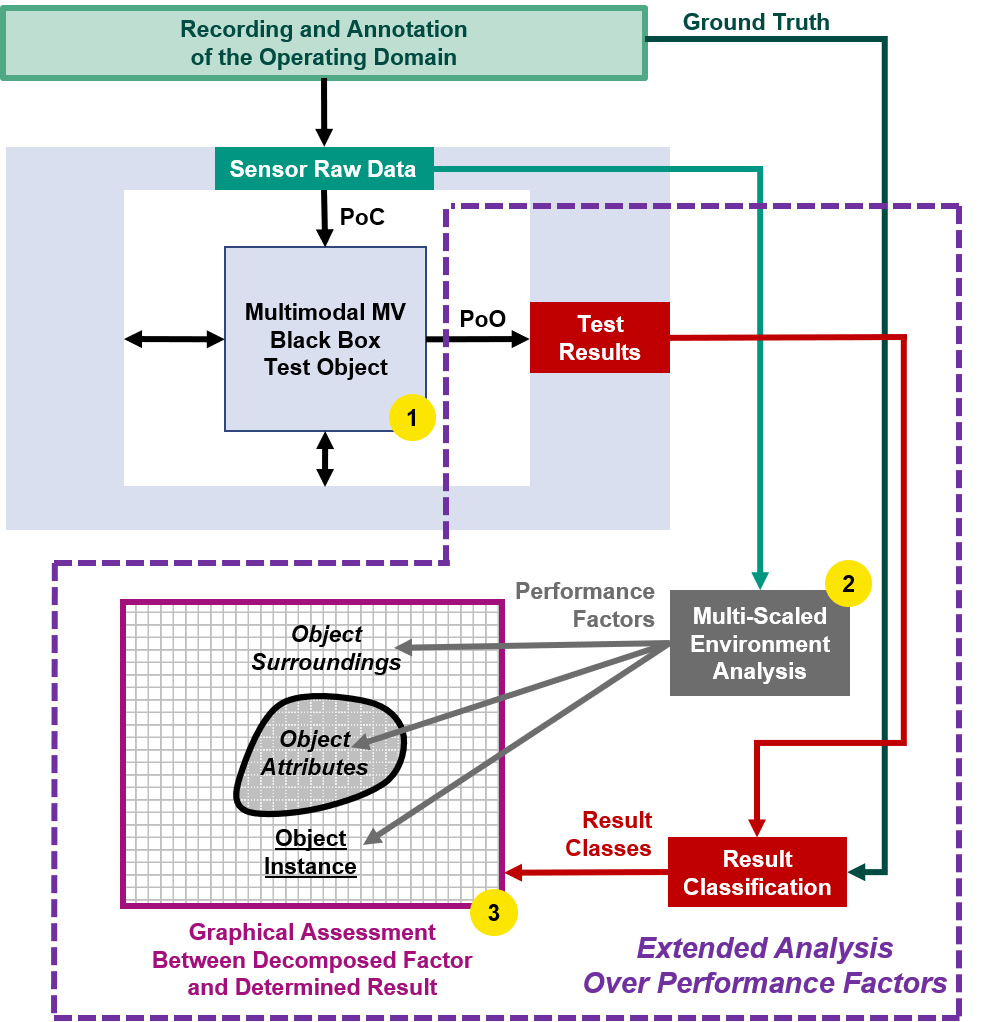}
        \end{adjustbox}
            \caption{General structure of the black-box testing framework incorporating selected solution from all three MV strategies by extending the analysis on performance factors relevant for MV with focus on environmental relations}
            \label{fig:general_procedure}
\end{figure}

Our collaborative strategy aims to combine and systematize selected solutions and the knowledge of the three MV strategies to develop a test method that is suitable for MV in development processes and homologation. Fig. \ref{fig:general_procedure} shows a general overview of our concept of a dynamic testing approach with an extended analysis of the points of observation in to obtain sufficient robustness statements. In this paper, the data provision and the systematic variation of the required stimuli is initially neglected. The focus lies on what scope needs to be considered for a holistic argumentation for MV test objects.
\begin{enumerate}
    \item To solve the time constraints in development processes described in challenge 1, we aim to evaluate multimodal MV test objects as a black box. This means that conclusions can only be drawn based on the inputs and outputs of the test object. This requires a discussion of which Points of Observation (PoO) are required to analyze the general concerns about MV. In sec. \ref{sec:crd}, we took a retrospective examination of the knowledge uncovered from XAI and CCA and the factors that led to these MV-specific performance losses.
    \item We apply a multi-scaled environment analysis to the input data, the sensor raw data, to divide the environment into individual possible MV-influencing entities. This approach supports us to overcome challenge 2 by relating performance not only to the targeted object instances, but also on their attributes and their immediate surroundings. This effective and efficient gradual structuring of the environment has led us to create the taxonomy of \textit{Granularity Orders}, described in sec. \ref{sec:go} 
    \item By identifying the MV deficits and their factors, we can transfer these previous blind spots to SBT and thus address challenge 3. SBT requires stating performances to associated relations of circumstances across a variety of driving scenarios. We do this by classifying the results against the ground truth for each environmental entity. Afterwards, the individual entities extracted from the granularization of the environment must be related to each other into a graphical representation. This enables arguing deficits via visible recurring relations that lead to failed results. To visualize the scope of assessments, the concept provide an entity relations graph (sec. \ref{sec:eerg}) and tested its applicability (ch. \ref{sec:application}).
\end{enumerate}
\begin{table*}[h]
    \vspace{2mm}
    \footnotesize
    \centering
    \caption{Identifications of General Object Recognition Deficits as Fundamental Concerns over MV Systems}
    \begin{tabularx}{\textwidth}{XXXXX}
        \hline
        \textbf{Deficits} & \textbf{Psychological Deficit} & \textbf{Psychological Identifications} & \textbf{Technical Translation} & \textbf{Technical Example} \\
        \hline
        \textit{Incomplete Domain Knowledge} & Connection to Semantic Knowledge & Everyday objects appear foreign and unfamiliar & Agent cannot recognize required objects generally in the target domain & Agent cannot recognize persons in images generally in targeted forest domain \\
        \hline
        \textit{Inapplicable Foreground-Background Differentiation} & Form-Ground Differentiation & Separation of a pattern into a figure and the background is not possible & Agent cannot recognize required objects under a specific scene & Agent recognize people in front of buildings but fails in front of park landscapes \\
        \hline
        \textit{Inapplicable Foreground-Foreground Differentiation} & Simultaneous Agnosia & Lack of integration of separate object parts & Agent cannot recognize required objects when specific object groups occur simultaneously & Agent is unable to segment or recognize a person from a crowd \\
        \hline
        \textit{Incomplete Shape Representation} & Form Agnosia & Incorrect recognition of curvatures and surface extensions & Agent cannot recognize required objects based on their shape appearing in sensor data & Agent does not recognize people if they are under 1.70 m \\
        \hline
        \textit{Incomplete Rotary Representation} & Insufficient Transformation into Object-centered Representation & Lack of mental rotation of object views & Agent cannot recognize required objects from specific perspectives & Agent recognizes people from their front view but not from their side view \\
        \hline
        \textit{Missing Attribute Integration} & Deficiency in Integrating Local Components & Figure is analyzed point by point & Agent cannot associate separately occurring attributes of a required object & Agent sees only one arm of a person sticking out and does not recognize that a human is there \\
        \hline
        \textit{Faulty Pattern Association} & Access to Object Recognition Unit & Inapplicable discrimination between real and art objects & Agent cannot correctly discriminate patterns by their purpose &  Agent recognizes the pattern of a T-shirt print as a traffic sign \\
        \hline
    \end{tabularx}

\label{tab:deficit_table}
\end{table*}
\subsection{Common Recognition Deficits}
\label{sec:crd}

Over the last decade, deep dive strategies have provided valuable insights into the exclusive behaviours and fundamental concerns of neural network functions. They provide information on which deficits need to be taken into account and which factors lead to these deficits. A review of these causes and effects shows what needs to be considered in a evaluation of MV. To get a guide for finding a pattern of general deficits, we made a comparison with human object recognition deficits and transferred their similarities to MV. The term "perception" finds its roots in psychological research, where the focus lies on visually capturing and converting form structures into meaningful semantics through neural processing chains in brains. Using insights from the literature of human perceptual psychology \cite{leschnik2020visuelle} \cite{frisby_stone_2010} \cite{karnath_thier_2012}, seven common human recognition deficits were gathered. These analogous dysfunctions in neural networks have been identified in publications and summarize them like presented in Table \ref{tab:deficit_table}:

\begin{enumerate}
    \item \textbf{Incomplete Domain Knowledge:} 
        Incomplete trained knowledge over usual relations in an operating domain. Originated from imbalanced or incomplete training data for MV function \cite{oksuz2020imbalance}
    \item \textbf{Inapplicable Foreground-Background Differentiation:} 
        Difficulty in recognizing a foreground entity due to issues with the simultaneous scene or background. Challenges may arise from foreignness, scene irritation, or entity adaptation to the background \cite{xiao2020noise}\cite{ribiero2016}. 

    \item \textbf{Inapplicable Foreground-Foreground Differentiation:} 
        Difficulty in separating a foreground entity when dealing with simultaneous equal or unequal groups of entities. Challenges may arise from foreignness, group irritation, or entity adaptation into or blockage by the group\cite{dollar2012}\cite{Cho2010UnsupervisedDA}.

    \item \textbf{Incomplete Object Representation:} 
        Deficits in object recognition explicitly related to the object instance itself, leading to nonrecognition or false classifications. Differences in holistic attributes like size, position, or pose making individuals unrecognizable. Discrepancies in size, shape, color, position, and gender can pose challenges for imbalanced MV \cite{Chan_2023}\cite{Rubinstein_2013_CVPR}.
    
    \item \textbf{Incomplete Rotary Representation:} 
        Difficulties caused by uncertainties on perspectives of specific instances. The neural network model may be over-optimized on certain views \cite{kanezaki2018rotationnet}\cite{Yu20213DOR}.

    \item \textbf{Object Component Integration into Object Knowledge:} 
        Challenges in understanding the relationship between a local element of an object's attributes and its parent class \cite{Xie_2021_ICCV}\cite{lindenberger2023lightglue}.

    \item \textbf{Faulty Pattern Association:} 
        Literals, symbols, shadows, and sensor noise artifacts can cause confusion for MV, especially if they don't appear or deviate in training data. Adversarial attacks with patterns can also lead to failures \cite{hassaballah2015face}\cite{thys2019fooling}. 
\end{enumerate}

By reflecting the identifications, we realize that humans and MV can have similar deficits, but the causes and underlying effects differ due to their different information processing mechanisms. An illustrative example is that drivers may experience trouble by encountering persons wearing dark clothing in darken environments. In contrast, MV might misclassify the cars potentially with a completely  not associated class. In Summary, the deficits not only show that recognition performance is allocated on properties of individual instances, but also on a multitude scaling, like attributes of different object or simultaneous surroundings, like object group or even the whole scenery. Also each factor does not stand alone as the relations between different granulated factors, like individual instances behind specific backgrounds can be a factor arising a failure. 

\subsection{Definition of Granularity Orders}
\label{sec:go}

\begin{figure}[h]
    \centering
      \begin{adjustbox}{margin=4ex 4ex,width=0.5\textwidth}
        \includegraphics[width=0.8\textwidth]{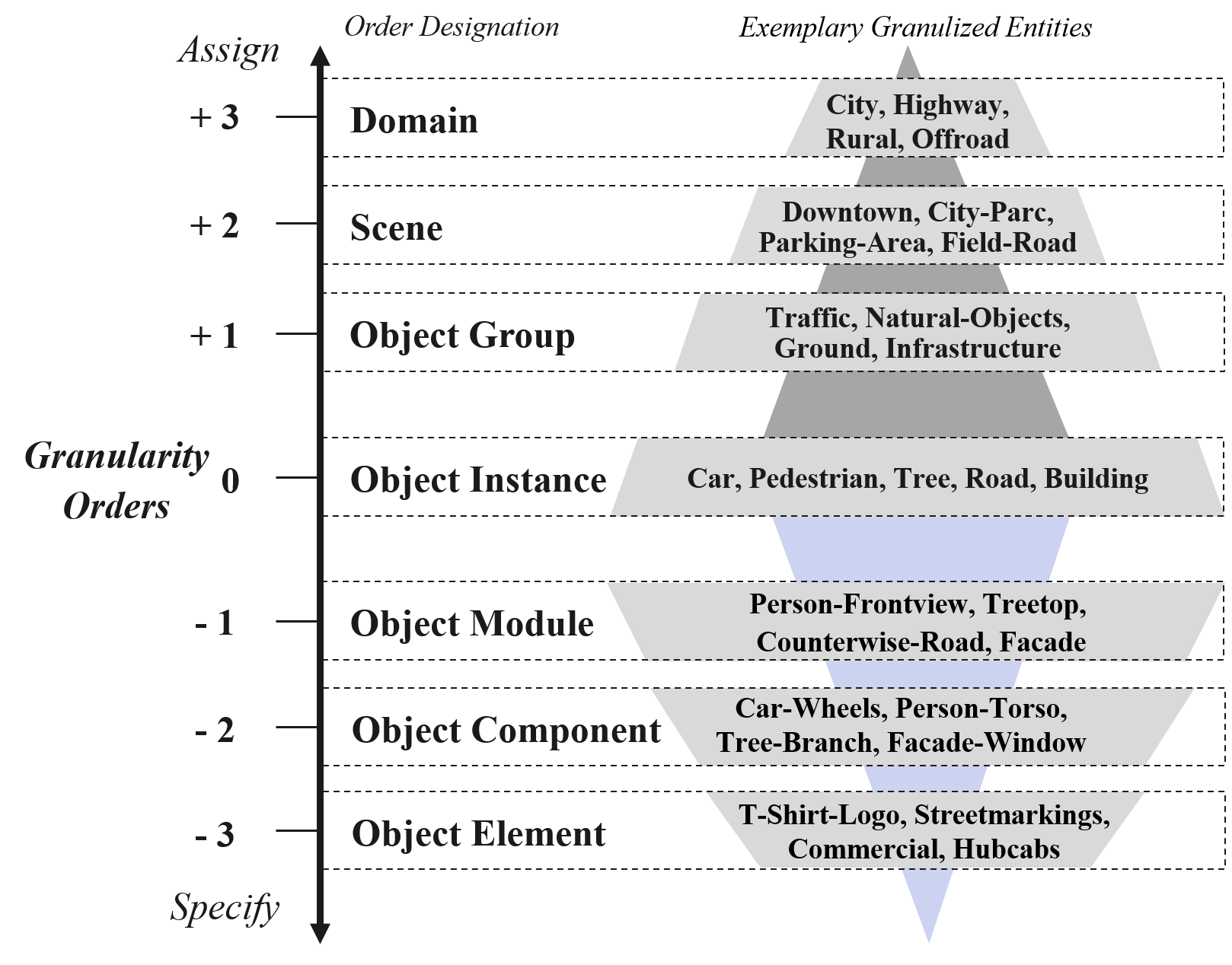}
        \end{adjustbox}
            \caption{Taxonomy of granularity orders structuring explored performance factors of ODD's into seven orders based on their informational depth}
            \label{fig:granularity_orders}
\end{figure}
The recurrent pattern of information for each deficit has led us to divide the operational design domain into seven order. Each of these orders ensures to explore the required performance factors to argue the deficits. Since this pattern regularly supported us to decipher unexpected results during scenario based test executions, we decided to formalize the orders into a taxonomy. This taxonomy divides the complexity of the ODD into increasingly deeper layers. While the 6-layer model \cite{scholtes21} expands across the breadth of the operational design domains and structures the variety of objects, the \textit{granularity orders} structures its elevation by recognizing their informational depth. Fig. \ref{fig:granularity_orders} gives an overview on all orders and each an example entity:

\textbf{Order 0 (Object Instance):} At the center order, all explored types of single instances are declared in the understanding of labels in panoptic segmentation. Dynamic objects, such as 'car', contrast with static objects, like 'building', and infrastructural objects, such as 'traffic sign' or 'road'. Although natural objects like 'tree' and 'sky' are single instances in an environment.

\textbf{Order -1 (Object Module):} This first specifying order, declares the rotative perspective of explored object instances from the MV perspective. An object should be declared by its view, such as front, back, or side. Improved semantics for order -1 describe familiar associations, such as the visibility of a building's facade from a MV perspective.

\textbf{Order -2 (Object Component)}: For the MV visible perspective of the explored instances the essential components should be declared in order -2. Essential means the characterizing parts of an object, where one part alone can give an association to the whole instance. Usually a "hand" is relatable to a "pedestrian".

\textbf{Order -3 (Object Elements):} The most granular order is exclusive to describe features, patterns and appearances on the surface of object components. Therefore entities that features letterings, patterns, curvature and reflections are allocated here.

\textbf{Order +1 (Object Group):} In this order, all instances are assigned to common groups based on their similarities in appearance. The naturalness, dynamics, and size of the instances serve as an indication of their similarity. The entities in order +1 are assigned to summary label categories that are declared in semantic segmentation datasets, such as vegetation, trees, and bushes, which can be assigned to natural objects.

\textbf{Order +2 (Scene):} In this orders the accumulation of all objects of a time frame results in a scene. The circumstances of a dominant occurrence of natural object groups, as also downtown-buildings and city-props can be assigned to city-park.

\textbf{Order +3 (Domain):} The highest order focuses on providing an overarching description of the operational domain. This should indicate to the geographical characteristics and context in which all scenes can be assigned to common category. For example, a "downtown", "suburb" and "city-park"-scenes can be assigned to the category "city".

The model requires the organisation of object instances by the user, whereby their attributes must be specified and they must be assigned to their environment. This helps to overcome challenge 2 of the generalisation strategy, as statements by averaging are only argumentatively sufficient over instances, but not under the circumstances that arise. Instead of averaging performance metrics, the use of granularity orders can reveal the exact causes for the occurrence of failures. The technical hurdle is that these multi-scaled factors must be generated for each individual data set. In the long term, this cannot be realised with terabytes or petabytes of recorded sensor data using a proper manual labelling process. Therefore, semi-automatic or fully automatic labelling is required for efficient application. On the other hand, a fully valid extraction of all these attributes and surroundings are necessary, as we want to make statements about high-risk systems. 

\subsection{Environmental Entity Relation Graph }
\label{sec:eerg}
\begin{figure}
    \centering
    \vspace{3mm}
    \includegraphics[width=0.99\linewidth]{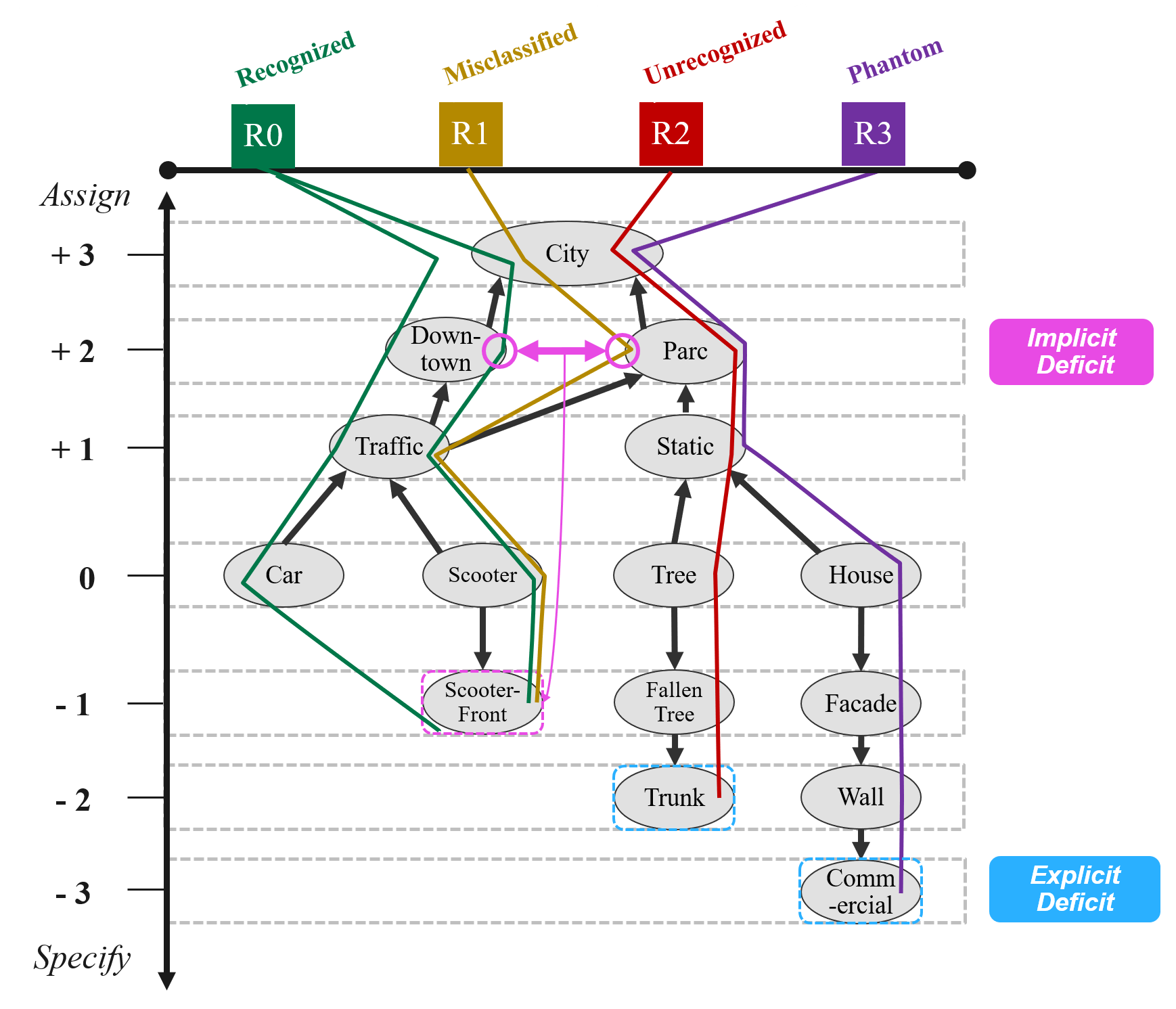}
    \caption{Schematic of the Environmental Entity Relation Graph}
    \label{fig:granularity_systematic}
\end{figure}

From a methodological point of view, granularization allows more precise statements to be made on explicit failures from MV. On the other hand, it should be noted that not all of the deficits listed can be explained by the occurrence of individual factors alone. Some deficits arise from relations between factors at various granularity: an inapplicable foreground-background differentiation can arise from a relation between an entity, like person at order 0 to a certain scene, like forest at order +2. To visualize this kind relations, a graphical representation was developed. The schema of the Environmental Entity Relation Graph (EERG) in Fig. \ref{fig:granularity_systematic} shows the relation graph network and its classified relations. A schema of the network incorporating these environmental entities is visible as ellipses and arrows.

\textbf{Relation Graph Network:} Each environmental entity has a relation to other entities of the composition ODD. A specific object instance can be assigned to its superordinate and be specified to its subordinate entities are depicted through arrow connections. A single chain from the highest to the lowest order, is a representation of an \textit{environmental relation}. Exemplary environmental relation: \textit{City-Parc-Static-House-Facade-Wall-Commercial}

\textbf{Classified Relation:} To argue about all deficits, each environmental relation must be classified according to the detected results of the test object. A classified relation is a line starting from the box of the result class and stretches down until it reaches each explicit entity. To precisely classify results we have defined one passed and three failed result classes:
    \begin{itemize}
    \item[-] \textbf{R0 Recognized:} A passed result that matches a tolerable deviation and the class with the ground truth.
    \item[-] \textbf{R1 Misclassified:} A failed result that matches a tolerable deviation but not the class with the ground truth.
    \item[-] \textbf{R2 Unrecognized:} A failed result that does not match a tolerable deviation with the ground truth but matches the class.
    \item[-] \textbf{R3 Phantom:} A failed result that does not match a tolerable deviation with any of the results of the ground truth within a simultaneous time frame.
    \end{itemize}
\textbf{Explicit Deficit:}
These Deficits are identifiable when failures are identifiable in a single time-frame on sensor-data compared to ground-truth. The schema shows two: (1) A commercial causing a faulty pattern association; (2) A not recognized lying trunk with a missing attribute integration to a fallen tree.

\textbf{Implicit Deficit:}
These Deficits are only identifiable over multiple time-frames with various driving conditions. They are visually recognisable by comparing neighbouring relations and according to ambivalent results that point to deviations in higher orders. The schema shows that in one frame the scooter is recognized in downtown, but in another time frame was misclassified in parc. This identification supports to argue a foreground-background differentiation.

\textbf{Advantages:} The reasoning of system behavior over time frames and runs is essential for scenario-based testing. Since multi-scale environment analysis and evaluation with the EERG is only dependent on the usual sensor raw data, it can be used across test instances and MV setups. Thus, it can be used seamlessly in all integration steps of development where MV or its functions are part of the test object.

\vspace{-2pt}
\section{Application}
\label{sec:application}

\begin{figure*}[!htbp]
    \centering
    \includegraphics[width=0.8\linewidth]{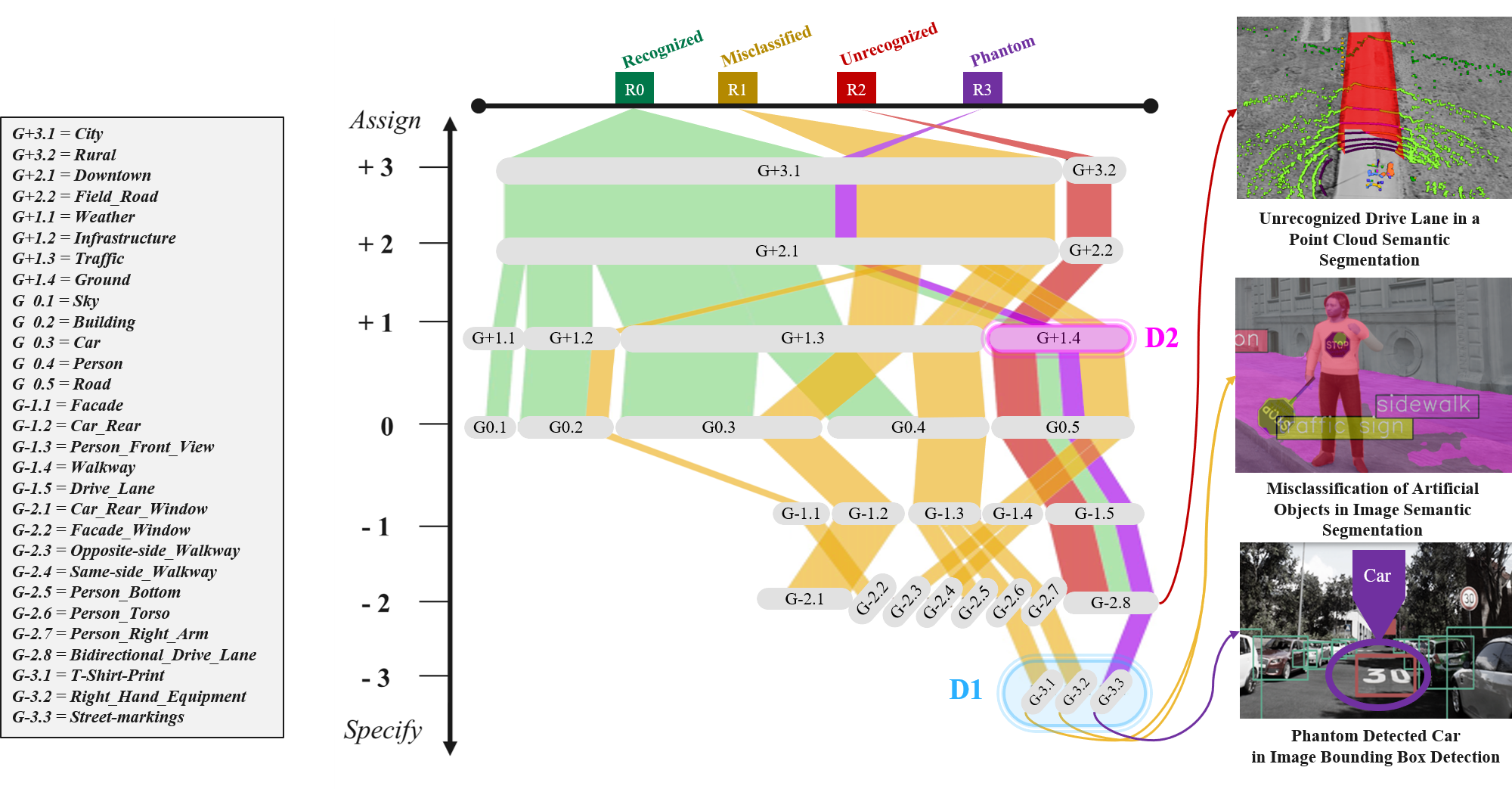}
    \caption{Exemplary Application of the Environmental Entity Relation Graph}
    \label{fig:results_classgraph}
\end{figure*}

This Application aims to reflect the visualization capabilities of the EERG and potential usage for detecting domain shifts or unexpected results. We analyze if deficits are arguable over a MV as black box, by this multitude and amount of assessment. To expose potential weaknesses arising from insufficient training data, we conducted two analyses. First, we analyzed the training data of three pre-trained models (YolovX, Deeplabv3, Cylinder3d) \cite{ge2021yolox} \cite{chen2017rethinking} \cite{zhu2021cylindrical} to identify potential limitations in the sensor data coverage for real-world scenarios. Second, we ran inferences on these models using data incorporating foreign entities. Our manual multi-scaled environment analysis led us to extract the granularized entities of three scenarios, which we listed and structured into the relation graph in Fig. \ref{fig:results_classgraph}. Three pictures highlight the most significant failures to declare two deficits.

\subsection{Identification of D1}

Of particular relevance is the unexpected phantom detection (violet) of the 30 km/h zone sign on the road as entity "G-3.3". This illustrates the limitations of scenario-based tests: It should be tested whether the MV test object is able to recognise a person crossing the road. However, the scenario failed repeatedly. In the first few seconds, the road marking was continuously detected with a bounding box of the class "car", which caused the driving system to stop. This is a serious blind spot in SBT, as statements about an entire test campaign on kinematic circumstances are falsified and effort is lost. Since the application of granularity orders recognises MV confounding factors, they support the minimisation of misinterpretation of test statements about scenarios.

During a manual inspection of the COCO dataset used to train the semantic segmentation, it was found that the images of the recorded labelled stop signs were predominantly upward-facing. In addition, some artificial stop signs were also labelled as real traffic signs. The Image segmentation was provoked in the simulation by the presentation of an image showing a person wearing a T-shirt with a stop sign logo and holding a lowered protest sign with the slogan "Stop Climate Change" (yellow). The image segmentation reacted explicitly deficient and classified the object elements as traffic signs. Both relations were classified as "misclassified". The granularity orders were also here able to recognise patterns that are not currently taken into account in a scenario-based test executions.

\textit{One way of minimising the deficit is to vary the spatially correct location of traffic signs and symbols in the domain. In addition, the checking of incorrectly labelled training data.}

\subsection{Identification of D2}
In an examination of the training and test data of the SemanticKITTI dataset \cite{Behley2019iccv}, results that the roadside in recorded parts of the city of Karlsruhe is dominated by pedestrian zones, trees and buildings. Only a few scenes consist of dominantly plain grounds. To find possible uncertainties in the lidar detection, we recorded point clouds also in Karlsruhe, but on a geometrically flat meadow scene. The EERG presents that the semantic segmentation was able to detect the environmental relation between specific roads for scenes with buildings at the roadside (the R1 classified bidirectional drive lane in “Downtown”). On the other hand, the model was not able to do so for the relation between similar roads to the meadow scenes (the R2 classified bidirectional drive lane in “Rural”). In the picture of the meadow scene is the explicit deficit shown (red), the model declares the road at a range more than 5 Meters as "Vegetation". The graph shows the deviation between on the environmental relation caused by changes of underrepresented surroundings. This shows a potential weakness in distinguishing between foreground objects, especially surfaces that need to be separated in forbidden and driveable areas, which appear to be geometrically similar. As in \textit{G+2.1} are occurring ambivalent classified results, this could indicate to an inapplicable foreground-foreground-differentiation, especially for ground objects.
\textit{An improvement of this uncertainty can be the saturation of objects that are geometrically similar in the domain but differ from the class in the training data.}

\section{Conclusion}
\label{sec:conclusion}
This paper highlighted the limitations of current testing strategies and propose a systematical handling with recognition deficits and their factors. Our novel framework concept decomposes ODDs into classifiable environmental relations acknowledging MV factoring environmental entities. Together with the new granularity oders taxnonomy and the EERG representation the framework enables the identification of MV uncertainties and minimize misinterpretations during test campaigns. It is noteworthy that this assessment is independent of system configurations and test platforms and can be used for applications, like detection of intolerable domain shifts.
In future works we want incrementally automatize the framework concepts and evaluate them on a sufficient amounts of data. For a useful application, the automation of the multi-scaled environment analysis on sensor data is elementary. This requires confident extraction of environmental entities from sensor data. Furthermore, there are open points about the generation of the input data with ground truth for a controlled stimulation during black box testing. Exemplary, the relation between foreground and background in stimulation data needs to be controlled for specific domains. 
The granularity orders and their application answer the question of sufficient depth, of the factors to be considered in the evaluation of MV: This can range from the entirety of a scene type to patterns that are derived from individual instances of objects. This selected depth allows us to identify relevant hurdles for the integration of MV, while avoiding the risk of running out of air. The sheer number of performance factors that need to be considered remains a major challenge. By applying the granularity orders, it is possible to focus on those most relevant to MV and thus enable targeted processing of the ODD information.

\section{Acknowledgement}
\label{sec:ack}
This work was supported by the German Federal Ministry for Economic Affairs and Climate Action within the project \textit{RepliCar} with grant number 19A23002I.

\bibliographystyle{IEEEtran}
\bibliography{root}
\end{document}